\documentclass{article}

\usepackage{microtype}
\usepackage{graphicx}
\usepackage{subfigure}
\usepackage{booktabs} 
\usepackage{amsmath,amssymb}
\usepackage{stmaryrd}
\usepackage{latexsym}
\usepackage{CJK}
\usepackage{indentfirst}
\usepackage{listings}
\usepackage{xcolor}
\usepackage{stmaryrd}
\usepackage{multirow}
\usepackage{verbatim}
\usepackage{makecell}
\usepackage{lipsum}
\usepackage[justification=raggedright]{caption}

\usepackage{hyperref}

\usepackage[accepted]{icml2021New}

\begin{document}

\twocolumn[
\icmltitle{Structured Model Pruning of Convolutional Networks on Tensor Processing Units}




\begin{icmlauthorlist}
\icmlauthor{Kongtao Chen}{g}
\icmlauthor{Ken Franko}{g}
\icmlauthor{Ruoxin Sang}{g}
\end{icmlauthorlist}

\icmlaffiliation{g}{Google, Mountain View, California, USA}

\icmlcorrespondingauthor{Kongtao Chen}{kongtao@google.com}

\icmlkeywords{Model Pruning, Tensor Processing Units}

\vskip 0.3in
]



\printAffiliationsAndNotice{}  

\begin{abstract}
The deployment of convolutional neural networks is often hindered by high computational and storage requirements.
Structured model pruning is a promising approach to alleviate these requirements.
Using the VGG-16 model as an example, we measure the accuracy-efficiency trade-off for various structured model pruning methods and datasets (CIFAR-10 and ImageNet) on Tensor Processing Units (TPUs).
To measure the actual performance of models, we develop a structured model pruning library for TensorFlow2 to modify models in place (instead of adding mask layers).
We show that structured model pruning can significantly improve model memory usage and speed on TPUs without losing accuracy, especially for small datasets (e.g., CIFAR-10).

\end{abstract}

\section{Introduction}
\begin{figure*}[!h]
\centering
\includegraphics[width=\textwidth]{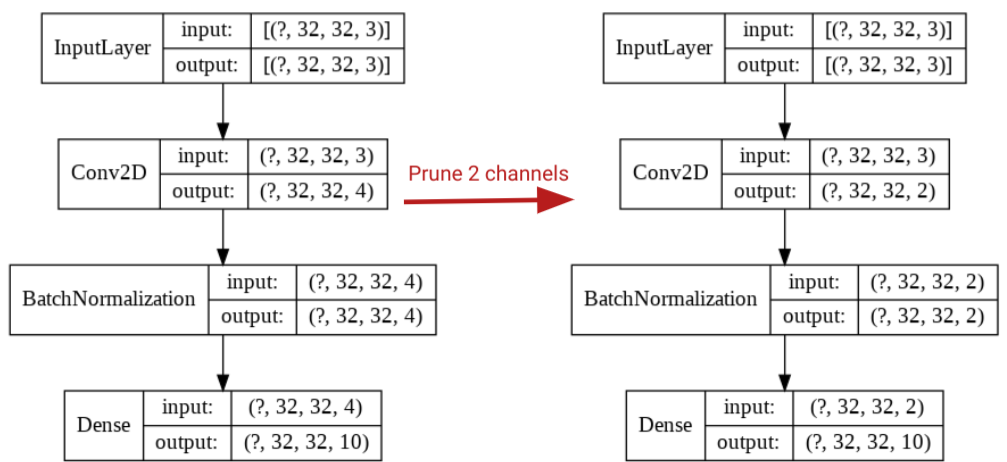}
\caption{\label{Prune} Illustration of the channel-based structured model pruning. 
Two channels from a Conv2D layer are pruned.}
\end{figure*}

Convolutional neural networks (CNNs) is the dominant approach for many computer vision applications, e.g., image classification \cite{Krizhevsky2012}, object detection \cite{Girshick2014}, semantic segmentation \cite{Long2015}.
The deployment of CNNs in real-world applications, however, is often constrained by model size, memory usage, and computational time.
Model pruning is one approach to compress model sizes and accelerate inference \cite{Liu2015,Han2016,Wen2016,Zhou2016,Li2017,Scardapane2017,Anwar2017,Zhang2018,Zhu2018,Deng2020}.
Model pruning can be realized at different levels, e.g., weight-level, channel-level, or layer-level. 
Weight-level (unstructured) pruning has the highest flexibility but usually requires special software or
hardware for fast inference \cite{Han2016}. 
Layer-level structured pruning does not require special packages for inference acceleration but is only effective when the depth is sufficiently large \cite{Wen2016}.
Channel-level structured pruning (shown in Fig. \ref{Prune}) balances flexibility and ease of implementation on general accelerators \cite{Liu2017,Li2017}.

Tensor Processing Units (TPUs) are the dominant approach deep learning accelerators for Google \cite{Jouppi2018,Jouppi2020}.
TPUs can be an order faster and more energy-efficient than contemporary GPUs or CPUs \cite{Jouppi2017}. 
At Google, an auto-tuner for the Accelerated Linear Algebra (XLA) compiler was developed to search for the fastest fusion configurations of an XLA program on TPUs \cite{Kaufman2020}.
Unstructured model pruning methods won't improve efficiency on TPUs because the XLA layout algorithm will pad zeros to the removed individual weights \cite{Kaufman2020}.
Whether structured model pruning can lead to efficiency gain on Google TPUs with XLA auto-tuners, however, remains unclear.

In this work, we measure the accuracy-efficiency trade-off for various channel-level structured model pruning methods and datasets on Google TPUs with XLA auto-tuners.
Performance (accuracy, memory usage and step time) of pruned models are measured on Google Borg \cite{Verma2015}.

\section{Methods}
We use a TensorFlow 2 \cite{Abadi2016} implementation of VGG-16 \cite{Simonyan2015} trained on ImageNet \cite{Russakovsky2015} and CIFAR-10 \cite{Krizhevsky2009} as an example to study the accuracy-efficiency tradeoff of structured model pruning on TPUs.
We train the original model, prune a certain ratio of channels, train the pruned models, and then measure model performance (accuracy, memory usage and step time) during the last epoch of training on Google Borg \cite{Verma2015}.

To find channels to prune, we use scaling factor-based \cite{Liu2017} and L1 norm-based pruning algorithms \cite{Li2017}.
These algorithms estimate the "importance" of a channel based on weights after training, e.g., $\gamma$ coefficient in the following batch-normalization layer (scaling factor \cite{Liu2017}) or L1 norm of channel weights (L1 norm \cite{Li2017}).
The least "important" channels will be pruned.
We also try whether reloading weights from the original model helps training of pruned models.
To create the pruned models without additional mask layers (so that the actual memory usage and step time can be measured), we create new models with desired structure and then reload corresponding weights from original models if desired.

For all trainings (including original models and pruned models), we use the same hyper-parameters (batch size $128$, max epochs $100$) and optimizers (SGD \cite{Bottou2010} with momentum $0.9$ and learning rate $0.01$).
This is because we aim at the accuracy-efficiency tradeoff of structured model pruning on TPUs rather than the best accuracy. 

\section{Results}

\begin{figure*}[!h]
\centering
\includegraphics[width=\textwidth]{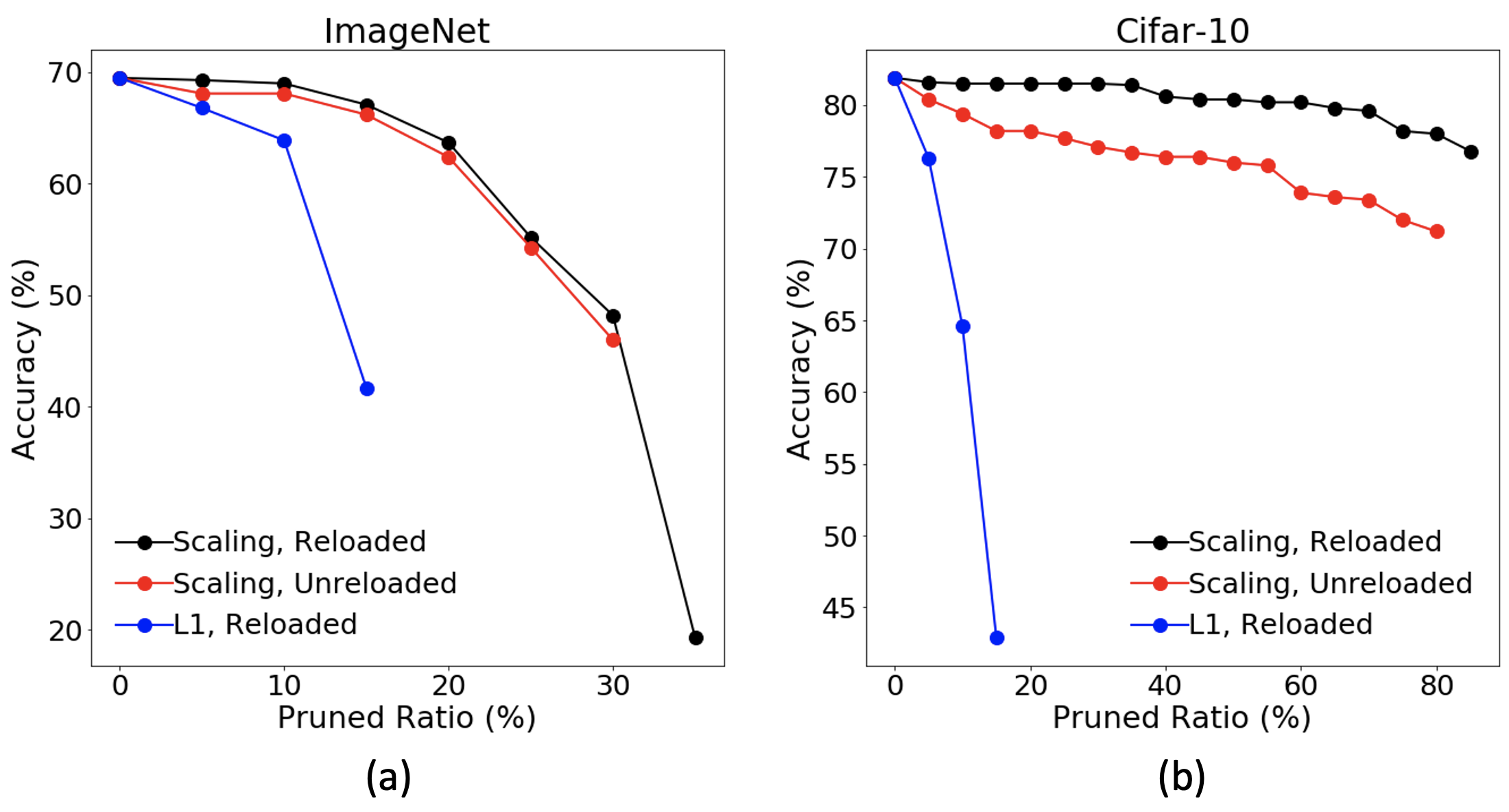}
\caption{\label{Accuracy} Accuracy as a function of channel pruned ratio on TPUs for CIFAR-10 and ImageNet for various pruning methods. 
Black curves use the scaling factor-based pruning algorithm \cite{Liu2017} and reload weights from original models for training. 
Red curves use the same pruning algorithm but don't reload weights from original models.
Blue curves use L1 norm-based pruning algorithm \cite{Li2017} and reload weights from original models.}
\end{figure*}

\begin{figure*}[!h]
\centering
\includegraphics[width=\textwidth]{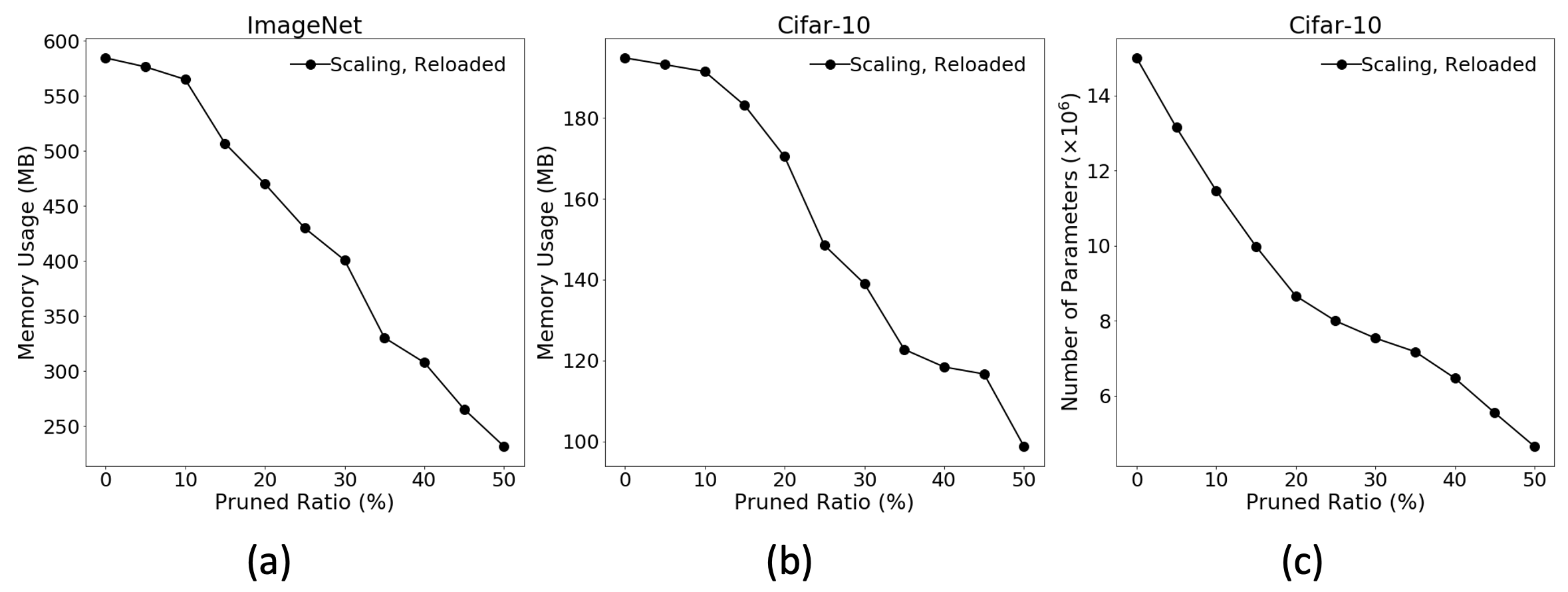}
\caption{\label{Memory} Memory usage and number of parameters as a function of channel pruned ratio on TPUs for CIFAR-10 and ImageNet.}
\end{figure*}

\begin{figure*}[!h]
\centering
\includegraphics[width=\textwidth]{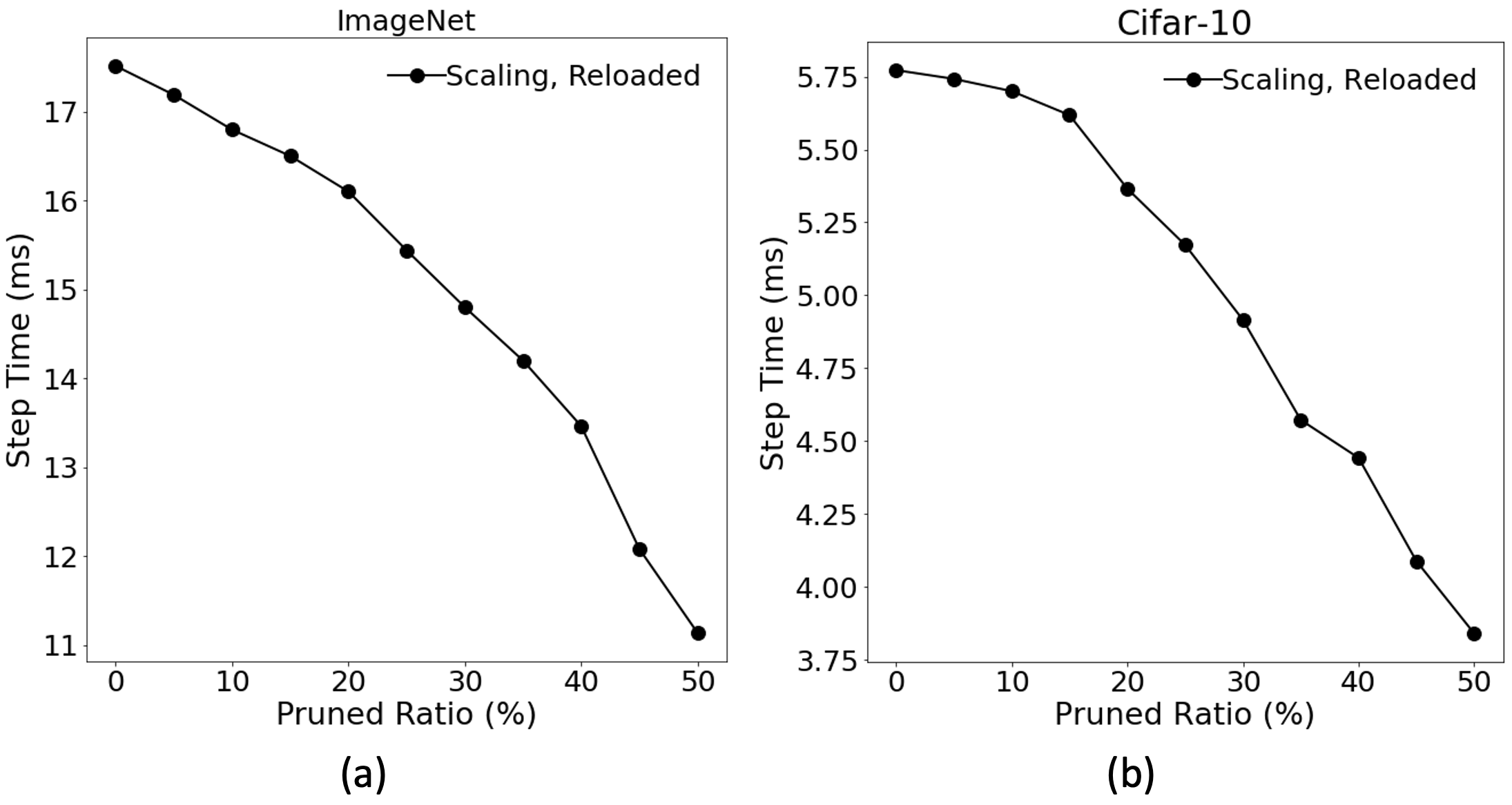}
\caption{\label{Time} Training step time as a function of channel pruned ratio on TPUs for CIFAR-10 and ImageNet.}
\end{figure*}

Figure \ref{Accuracy} shows accuracy as a function of the pruned ratio of channels on TPUs for CIFAR-10 and ImageNet for various pruning methods.
As pruned ratio increases, the accuracy for ImageNet decreases much faster than accuracy for CIFAR-10.
To sustain original accuracy, VGG-16 can be pruned by $30\%$ on CIFAR-10 while $10\%$ on ImageNet. 
At $30\%$ pruned ratio, accuracy for CIFAR-10 is almost the same, while accuracy for ImageNet decreases by $20\%$, indicating that VGG-16 is over-parameterized for CIFAR-10.
CIFAR-10 requires fewer parameters to fit accurately compared with ImageNet because the optimal model size scales with the dataset size \cite{Hestness2017}.
VGG-16 is more over-parameterized on CIFAR-10 and a larger pruned ratio can be used without losing accuracy.
Accuracy decreases faster and faster when pruned ratio increases (the absolute value of the slope increases) because the first parameters pruned are over-parameterizations.
For both datasets, reloading weights from the original model improves pruned accuracy, and scaling factor-based pruning \cite{Liu2017} leads to better accuracy than L1-norm-based pruning \cite{Li2017}.

Structured model pruning can improve training memory usage (Fig. \ref{Memory}) and step time (Fig. \ref{Time}) on TPUs.
While the saved memory usage, number of pruned parameters, saved step time, and pruned ratio of channels are proportional to each other, they are not strictly linear with each other.
This is because the number of parameters of each channel is different, and TPUs will pad zeros to weights according to the XLA layout algorithm \cite{Kaufman2020}.

\section{Conclusion}
Using the VGG-16 model as an example, we measure the accuracy-efficiency trade-off for various structured model pruning methods and datasets (CIFAR-10 and ImageNet) on TPUs.
In both cases, reloading weights from the original model improves pruned accuracy, and scaling factor-based pruning \cite{Liu2017} leads to better accuracy than L1-norm-based pruning \cite{Li2017}.
We show that structured model pruning can significantly improve model memory usage and speed on TPUs without losing accuracy, especially for small datasets (e.g., CIFAR-10).
This is because VGG-16 is over-parameterized for a small dataset like CIFAR-10.
Our results suggest that structured model pruning is a promising approach to improve the efficiency of CNNs on TPUs.

\clearpage

\bibliography{mybib}
\bibliographystyle{icml2021}

\end{document}